\documentclass{article}

\PassOptionsToPackage{numbers}{natbib}

    \usepackage[preprint]{neurips_2024}

\usepackage[utf8]{inputenc} 
\usepackage[T1]{fontenc}    
\usepackage{hyperref}       
\usepackage{url}            
\usepackage{booktabs}       
\usepackage{amsfonts}       
\usepackage{nicefrac}       
\usepackage{microtype}      
\usepackage{graphicx}
\usepackage{subcaption}
\usepackage{listings}
\usepackage{xcolor}         
\usepackage{cleveref}

\usepackage{enumitem}   
\usepackage{bm}
\usepackage{pifont}

\title{Leveraging Unstructured Text Data for Federated Instruction Tuning of Large Language Models}

\author{%
  Rui Ye\textsuperscript{1}\footnotemark[1] $\quad$ Rui Ge\textsuperscript{1}\footnotemark[1] $\quad$ Fengting Yuchi\textsuperscript{1} $\quad$
  \textbf{Jingyi Chai}\textsuperscript{1}  $\quad$
  \textbf{Yanfeng Wang\textsuperscript{2,1}} $\quad$ \textbf{Siheng Chen\textsuperscript{1,2}} \\\\
  \textsuperscript{1} Shanghai Jiao Tong University $\quad$ \textsuperscript{2} Shanghai AI Laboratory \\
}

\begin{document}

\maketitle
\renewcommand{\thefootnote}{\fnsymbol{footnote}}
\footnotetext[1]{Equal contribution.}

\begin{abstract}
  Federated instruction tuning enables multiple clients to collaboratively fine-tune a shared large language model (LLM) that can follow humans' instructions without directly sharing raw data.
  However, existing literature impractically requires that all the clients readily hold instruction-tuning data (i.e., structured instruction-response pairs), which necessitates massive human annotations since clients' data is usually unstructured text instead.
  Addressing this, we propose a novel and flexible framework FedIT-U2S, which can automatically transform unstructured corpus into structured data for federated instruction tuning.
  FedIT-U2S consists two key steps:
  (1) few-shot instruction-tuning data generation, where each unstructured data piece together with several examples is combined to prompt an LLM in generating an instruction-response pair.
  To further enhance the flexibility, a retrieval-based example selection technique is proposed, where the examples are automatically selected based on the relatedness between the client's data piece and example pool, bypassing the need of determining examples in advance.
  (2) A typical federated instruction tuning process based on the generated data.
  Overall, FedIT-U2S can be applied to diverse scenarios as long as the client holds valuable text corpus, broadening the application scope of federated instruction tuning.
  We conduct a series of experiments on three domains (medicine, knowledge, and math), showing that our proposed FedIT-U2S can consistently and significantly brings improvement over the base LLM.
\end{abstract}

\section{Introduction}

Instruction tuning has become one of the most imperative components in training contemporary instruction-followed large language models (LLMs)~\cite{openai2023gpt4,llama2,jiang2023mistral,yang2024qwen2}, where typically, the training samples are collected from diverse sources by a central party~\cite{ouyang2022training,wei2021finetuned,zhou2023lima}.
However, these data could contain sensitive (e.g., private or proprietary) information that cannot be directly shared, making such centralized learning paradigm inapplicable especially for domains such as medicine~\cite{singhal2023towards} and finance~\cite{wu2023bloomberggpt}.

Addressing this, federated learning~\cite{fedavg,kairouz2021advances} has emerged as a well-suited technique to achieve instruction tuning of LLMs without direct data sharing.
In federated instruction tuning (FedIT), each party (i.e., client) keeps its private data locally and shares the instruction-tuned LLM with the central server, while the server aggregates LLMs from multiple parties and distributes the aggregated LLM back to participating parties.
Such paradigm has attracted massive attention and interests from both academia~\cite{ye2024openfedllm,ye2024fedllm,fedit} and industry~\cite{fedml-fedllm,fan2023fate,federatedscopellm}.

Despite extensive efforts dedicated to FedIT, existing methods impractically rely on the assumption that each party possesses structured instruction-tuning data (i.e., instruction-response pairs), which significantly constrains the real-world applicability of FedIT.
In practice, while clients may possess valuable data locally, this data often exists in an unstructured format (just strings of text) rather than naturally aligns with the structured format required for IT~\cite{genie}.
Consequently, current FedIT systems face challenges in scalability, as they necessitate manual annotation of data by each client.

To fill this gap, we propose a novel and flexible framework FedIT-U2S, which can automatically transform unstructured corpus into structured instruction-tuning data for FedIT, bypassing the massive human efforts required for data annotation.
Specifically, FedIT-U2S consists of two key steps: few-shot instruction-tuning data generation and FedIT on the generated data.
(1) The server first distributes an open-sourced general LLM and a few examples (could be as few as only one) to participating clients.
During data generation, each client queries the LLM to generate multiple instruction pairs, where each pair is generated by feeding the LLM with a prompt that is composed of few examples as the context and a sampled piece of its unstructured data.
To further enhance the generality and scalability of FedIT-U2S, we propose a retrieval-based example selection approach, where for each sampled piece of unstructured data, similarity scores are computed by comparing it with all the examples sent from the server, after which the top-k examples are selected as the few-shot examples in the context for data generation.
(2) Subsequently, typical federated instruction tuning is launched based on the general LLM and the generated datasets in the previous step.
Considering communication and computation efficiency, LoRA~\cite{hu2021lora} is applied and therefore only a small set of parameters are learned and communicated.
Overall, our FedIT-U2S framework makes FedIT system as practical as Google's GBoard application (next word prediction)~\cite{hard2018federated}, where the supervision data directly comes from user's data without any manual effort.

To verify the effectiveness of our proposed framework, we conduct a series of experiments covering three domains (i.e., medicine, knowledge, and math).
We show that across these domains, our FedIT-U2S consistently improves the performance of the general LLM on the corresponding downstream task.
Besides, we show the effectiveness of several designs, including retrieval-based example selection and filtering during data generation, providing potential directions for further improving the performance of FedIT-U2S.

Our contributions are as follows:

\begin{enumerate}[leftmargin=*]
    \item We propose the first end-to-end framework (FedIT-U2S) for directly leveraging unstructured data for federated instruction tuning of large language models.
    \item We propose a retrieval-based example selection technique and a few-shot data generation mechanism, which automatically selects examples for higher relatedness and generates structured data in an expected manner.
    \item We verify the effectiveness of FedIT-U2S through a series of experiments on multiple domains.
\end{enumerate}

\section{Related Work}

\textbf{Federated Learning of Large Language Models.}
Federated learning is a privacy-preserving machine learning paradigm that enables multiple clients to collaboratively train machine learning models without sharing their raw data~\cite{fedavg,kairouz2021advances}.
With the rise of large language models (LLMs), researchers have recently begun to consider federated training of LLMs to safeguard client data privacy or to address the scarcity of publicly available data~\cite{villalobos2022will,ye2024openfedllm}, which has attracted massive attention and interests from both academia~\cite{ye2024openfedllm,ye2024fedllm,fedit} and industry~\cite{fedml-fedllm,fan2023fate,federatedscopellm}.

Specifically, OpenFedLLM~\cite{ye2024openfedllm} offers an integrated framework and provides a comprehensive empirical study to show the potential of federated instruction tuning of LLMs (FedIT).
Similarly, FederatedScope-LLM~\cite{federatedscopellm} and FedML-LLM~\cite{fedml-fedllm} provide frameworks that implement FedIT;
while FedLLM-Bench~\cite{ye2024fedllm} offers real-world datasets and benchmarks.
Besides frameworks and benchmarks~\cite{collins2023profit}, a series of methods are proposed to target various perspectives including safety alignment~\cite{ye2024emerging}, privacy~\cite{sun2024improving}, heterogeneous computation~\cite{cho2023heterogeneous}.

However, existing literature assumes that client data is structured in the form of instruction-response pairs, overlooking the reality that client data often exists in an unstructured format.
In such cases, clients are required to annotate data before participating in FedIT, which is labor-intensive and limits its broader adoption.
In this paper, we address this issue for the first time by proposing FedIT-U2S, a method that automates the transformation of unstructured client data into structured data prior to FedIT.
This reduces the need for manual annotation and broadens the applicability of FedIT.

\textbf{Data Generation in Large Language Models. }
The quality and quantity of data play a critical role in the training of large language models.
However, manually generating and annotating data is labor-intensive and hard to scale up.
Addressing this, the community turns to using LLMs to generate high-quality data~\cite{xu2023wizardlm,instruction-pretrain,adler2024nemotron,dubey2024llama}.
For example, Self-Instruct~\cite{wang2022self} leverages 8 in-context examples to prompt LLMs for generating new instruction samples.
WizardLM~\cite{xu2023wizardlm} instructs ChatGPT to generate diverse instructions via evolving prompt.
MATRIX~\cite{pangself} instructs the LLMs to generate data for value alignment via social simulation.
Genie~\cite{genie} employs few-shot methods~\cite{fewshot} to transform unstructured data into three kinds of structured data.
Instruction Pre-training~\cite{instruction-pretrain} generates instruction-tuning data to augment pre-training.

In this paper, we for the first time consider utilizing clients' unstructured data for FedIT of LLMs by leveraging the LLMs for data generation.
We apply few-shot generation technique for its simplicity and effectiveness; while we believe that there could be other techniques applied to our scenario.

\section{Methodology}

\begin{figure}[t]
    \centering
    \includegraphics[width=0.9\linewidth]{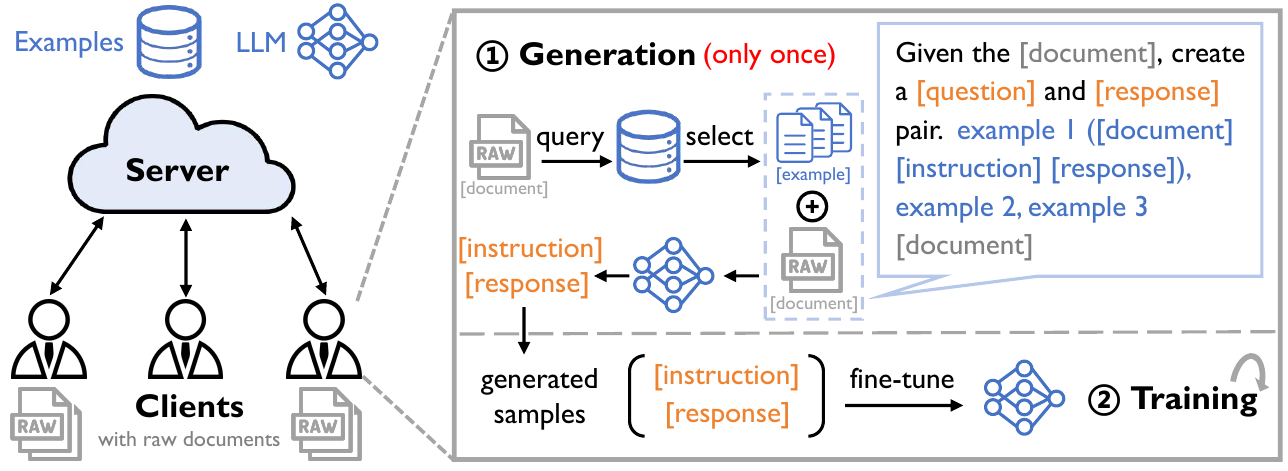}
    \caption{Overview of our proposed FedIT-U2S. It consists of two key steps: data generation and FedIT. Data generation is required only once before FedIT. (1) For each raw unstructured data piece, clients select a few examples by retrieving from an example database to construct a few-shot template, prompting the LLM to generate an instruction-response pair. (2) Typical federated instruction tuning starts based on the generated structured instruction-tuning data.}
    \label{fig:overview}
\end{figure}

In this section, we first introduce the overall framework of our proposed FedIT-U2S (Figure~\ref{fig:overview}), which consists of two key steps: few-shot instruction-tuning data generation (which transforms unstructured data into structured instruction-tuning data pairs) and federated instruction tuning on the generated data.
Then, we detail our design of retrieval-based example selection for few-shot data generation.

\subsection{Pipeline of FedIT-U2S}
\label{sec:pipeline}

At the beginning of FedIT-U2S, the server first distributes an open-sourced general LLM (denoted by $\bm{\theta}^*$) and a set of examples (unstructured and structured text pairs, denoted by $\mathcal{O}$) to participating clients.

\textbf{Step 1: few-shot instruction-tuning data generation.}
Suppose there are $M$ clients in the system and each client $m$ holds an unstructured dataset $\mathcal{D}^u_m=\{\bm{d}_i\}_{i=1}^{N_m}$, where $\bm{d}_i$ is a data piece and $N_m$ denotes the number of data pieces.
Since such unstructured data cannot be directly used for instruction tuning, it conventionally requires each client's efforts to manually create instruction-response pairs for tuning, which is costly and faces the challenges of scaling up.
To address this, we design to automatically transform the unstructured data into a structured instruction-response format via a few-shot data generation process, which leverages LLM's in-context learning capability~\cite{fewshot}.

Specifically, upon receiving example set $\mathcal{O}=\{(\bm{d}_i,\bm{x}_i, \bm{y}_i)\}_{i=1}^O$, where $O$ is the example number, $\bm{d}_i$ is an unstructured data document, $\bm{x}_i$ and $\bm{y}_i$ is the document-grounded instruction and response respectively, each client selects several (denoted by $k$) examples as few-shot examples prompt the LLM $\bm{\theta}^*$.
Denote the instruction for generation as $I$ and the selected examples as $\mathcal{S}=\{(\bm{\hat{d}}_i,\bm{\hat{x}}_i, \bm{\hat{y}}_i)\}_{i=1}^k$, given a user's data piece $\bm{d}$, the prompt $P$ is constructed as: $P = Concat (I, \mathcal{S}, \bm{d})$, where $Concat$ denotes the concatenation operation (see full prompt in~\ref{app}).
Note that these examples can be either randomly selected for diversity or selected according to relatedness between user data and examples for better diversity-relatedness trade-off, which will be detailed in Section~\ref{sec:retrieval}.
Based on the prompt, the LLM $\bm{\theta}^*$ will generate an instruction-response pair: $(\bm{x}, \bm{y}) = f (P;\bm{\theta}^*)$.
Therefore, by iterating on client's unstructured dataset $\mathcal{D}^u_m=\{\bm{d}_i\}_{i=1}^{N_m}$, we obtain a structured dataset for instruction tuning: $\mathcal{D}^s_m=\{\bm{x}_i, \bm{y}_i\}_{i=1}^{N_m}$.

Since the responses of LLMs are in an open-ended form and there are randomness during generation, some generated data might fall short in terms of data quality.
Therefore, additional data filtering is necessary for enhancing the data quality.
Here, we consider two filtering mechanisms: rule-based filtering to remove data with undesired format and reward-based filtering to ensure the quality of selected data.
Specifically, we first filter out data that does not follow the format of instruction-response pair.
Secondly, we use an publicly available reward model to score the generated data samples and select the top two-thirds samples.
This enables us to select data that is more aligned with human preference since reward model is trained to model human preference.

\textbf{Step 2: federated instruction tuning on the generated data.}
With the generated data, a typical process of federated instruction tuning is started.
Considering computation and communication efficiency, we apply LoRA~\cite{hu2021lora} as the parameter-efficient fine-tuning technique.
Suppose there are $T$ rounds of federated learning rounds in total.
At each round $t$, the server sends the model parameters $\bm{\theta}^{t}$ to each available client.
Then, each client $m$ initializes its local trainable parameters with $\bm{\theta}^{t}$, keeps the base model parameters $\bm{\theta}^*$ fixed, and starts supervised fine-tuning on its generated dataset $\mathcal{D}^s_m=\{\bm{x}_i, \bm{y}_i\}_{i=1}^{N_m}$, where the model learns to predict the response $\bm{y}_i$ given the instruction $\bm{x}_i$.
By fine-tuning for several steps, each client $m$ obtains a fine-tuned model parameters $\bm{\theta}_m^{t}$ and sends it to the server.
Finally, the server aggregates model parameters of clients to obtain the global model parameters for the next round: $\bm{\theta}^{t+1} = \sum_m p_m \bm{\theta}_m^{t}$, where $p_m=\frac{N_m}{\sum_i N_i}$ is the relative dataset size of client $m$.

\subsection{Retrieval-based Example Selection for Few-Shot Generation}
\label{sec:retrieval}

The chosen examples (i.e., the context) in the prompt could significantly affect the behaviour of LLMs~\cite{brown2020language,in-context}, resulting in different quality of the genrated data.
Therefore, to generate high-quality structured data, selecting appropriate few-shot examples is essential.
Generally, examples that closely match the target text in terms of content and structure tend to produce more effective results.
However, in practical applications, manually identifying suitable examples can be a time-consuming process, making it inflexible in adapting to diverse scenarios.
To mitigate this challenge, we propose a retrieval-based example selection method for few-shot generation which automatically selects few-shot examples from a mixed example pool according to similarity between user data and examples.

Given the set of examples sent from the server $\mathcal{O}=\{(\bm{d}_i,\bm{x}_i, \bm{y}_i)\}_{i=1}^O$, each client aims to select $k$ examples for each of its sampled unstructured data piece.
Specifically, for each data piece $\bm{d}$, we compute the similarity $Sim(\bm{d}, \bm{d}_i)$ for each $\bm{d}_i$ in the example pool $\mathcal{O}$ using BERT Score as the metric, which gives a similarity score that reflects the relatedness between the target data piece and the example's content.
Subsequently, we rank the similarity scores and select top-k examples $\mathcal{S}=\{(\bm{\hat{d}}_i,\bm{\hat{x}}_i, \bm{\hat{y}}_i)\}_{i=1}^k$, which are mostly likely to guide the LLM to generate high-quality and highly-related data.
The other procedures remain unchanged as in Section~\ref{sec:pipeline}.

\section{Experiments}

\subsection{Experimental Details}

\textbf{Training Dataset.}
We consider three datasets for our experiments~\cite{instruction-pretrain}, which cover domains including medicine, knowledge, and math.
Specifically, PubMedQA~\cite{pubmedqa} is a medical dataset for biomedical research question answering with corresponding abstracts as the context.
HotpotQA~\cite{hotpotqa} is a dataset of  Wikipedia-based questions with supporting facts as the context.
AQUA\_RAT~\cite{aqua-rat} is a math dataset for algebraic word problems answering. The problems, together with solutions, form the context.
We select 10,000 samples from each dataset for the experiments~\cite{instruction-pretrain}, with each sample comprising a piece of original unstructured text, along with a human annotated instruction and response, both derived from the text.
Only the unstructured text is used in our method FedIT-U2S, while the human annotated instruction-response pairs are used to implement FedAvg as a reference to verify the effectiveness of our method.

\textbf{Implementation Details.}
Our implementation is based on the open-sourced codebase OpenFedLLM\footnote{\href{https://github.com/rui-ye/OpenFedLLM}{https://github.com/rui-ye/OpenFedLLM}}~\cite{ye2024openfedllm}.
We use Vicuna-7B ~\cite{chiang2023vicuna} as the base model and set the learning rate as $2e^{-5}$ with a batch size of 16. The communication round is set to 200 and 2 clients are sampled out of 5 each round to participate federated instruction tuning.
We use \textit{reward-model-deberta-v3-large-v2} as the reward model following~\cite{genie}.
We select $k=3$ examples for few-shot generation.

\textbf{Evaluation Metrics.}
\textbf{(1) BERT Score:} BERT Score ~\cite{bertscore} is an evaluation metric for natural language generation that measures the similarity between a candidate sentence and reference sentences by leveraging contextual embeddings from pre-trained language models like BERT. 
\textbf{(2) ROUGE-L:} ROUGE-L~\cite{rouge} is an evaluation metric used for summarization and text generation tasks, focusing on the longest common subsequence (LCS) between a candidate sentence and a reference sentence. ROUGE-L evaluates the extent to which the candidate sentence preserves the order and content of the reference, providing a more holistic assessment of the generated text's quality.
We select 50 samples from each dataset to serve as the test set.
We compare the model-generated responses to the gold standard answers (i.e., human-annotated answers in the test set) by calculating BERT Score and ROUGE-L to assess performance.

\textbf{Compared LLMs.}
(1) The base LLM, i.e., the Vicuna model without additional tuning;
(2) the base LLM tuned via FedAvg on human-annotated data, which serves as a performance reference;
(3) the base LLM tuned by our FedIT-U2S without filtering technique;
and (4) the base LLM tuned by our FedIT-U2S with filtering technique.

\subsection{Experimental Results}

\textbf{Comparisons with baselines.}
In Table~\ref{tab:result}, we compare models trained via our methods on generated data with base model and model trained via FedAvg~\cite{fedavg} on human-annotated data (as a reference).
Experiments are conducted on three datasets and evaluated by two metrics.
From the table, we see that 
(1) our methods consistently and significantly improves the performance of the base model across datasets and evaluation metrics, indicating the effectiveness of our proposed methods.
Specifically, in HotpotQA, our method can achieve 0.1873 higher BERT Score (0.2439 v.s. 0.0566).
(2) Our methods hugely fill the gap between base model and that tuned via FedAvg on human data, further verifying FedIT-U2S's effectiveness.
However, there is still a room for improvement, calling for more future works to further enhance the performance.
With the increasing generation capability of LLMs~\cite{dubey2024llama,adler2024nemotron}, we even believe that there is potential for surpassing this baseline (FedAvg on human data).
(3) Although the data filtered using the reward model is smaller in quantity, it brings a more significant improvement to the model's performance, indicating the importance of data quality in this scenario.

\begin{table}[t]
\small
\setlength\tabcolsep{4pt}
\centering
\begin{tabular}{l|cc|cc|cc}
\toprule
& \multicolumn{2}{c|}{PubMedQA} & \multicolumn{2}{c|}{HotpotQA} & \multicolumn{2}{c}{AQUA\_RAT} \\
 & BERT Score & ROUGE-L & BERT Score & ROUGE-L & BERT Score & ROUGE-L\\
\midrule
Base Model & 0.1483 & 0.1496 & 0.0566 & 0.2380 & -0.0171 & 0.1529  \\
\textbf{FedIT-U2S} & 0.1876 & 0.1727 & 0.1774 & 0.2942 & 0.0885 & 0.2383  \\
\textbf{FedIT-U2S} (Filtered) & 0.2043 & 0.1859 & 0.2439 & 0.3226 & 0.1131 & 0.2452  \\
\bottomrule
FedAvg on Human Data & 0.2306 & 0.2017 & 0.2701 & 0.3531 & 0.1381 & 0.2890  \\
\bottomrule

\end{tabular}
\caption{Experiments on three datasets: PubMedQA (medical), HotpotQA (knowledge), and AQUA\_RAT (math). Our proposed FedIT-U2S consistently brings performnace improvement compared to base model. FedIT-U2S (Filtered) hugely fills the gap between base model and FedAvg on human-annotated data, indicating the effectiveness of our proposed method in bypassing massive human efforts in annotation.}
\label{tab:result}
\end{table}

\textbf{Analysis of example selection for few-shot generation.}
The effectiveness of few-shot generation may heavily rely on the chosen examples in the context.
Therefore, here, we deeply analyze the example selection by conducting a series of experiments on HotpotQA dataset since we observe a large improvement in previous experiments.
In this experiment, the example pool has 50 samples in total, covering five domains: medicine, math, knowledge, common sense, and daily life.
We consider the following setups of few-shot generation in our proposed FedIT-U2S:
\ding{172} Random 0 + 3: 3 out-domain examples are randomly selected (e.g., for medical task, examples from other domains are randomly selected);
\ding{173} Random 1 + 2: 1 in-domain and two out-domain examples are randomly selected;
\ding{174} Fixed 3 + 0: 3 fixed in-domain examples are selected for all generation;
\ding{175} Random 3 + 0: 3 in-domain examples are randomly selected;
\ding{176} Retrieval-based Selection: 3 examples are automatically selected from a mixed example pool by our retrieval-based example selection technique.

The experimental results are shown in Table~\ref{tab:domain}.
(1) Compared to the base model, \ding{172}, which introduces out-domain examples for few-shot generation, does not bring evident improvement while \ding{173}-\ding{176} all bring consistent improvement.
This indicates the importance of selecting appropriate examples for few-shot data generation.
(2) Comparing \ding{172}, \ding{173}, and \ding{175}, we can see that increasing the number of in-domain examples consistently brings more performance improvement, indicating the value of introducing in-domain examples to facilitate generation.
(3) Comparing \ding{174} and \ding{175}, we see that randomly selecting in-domain examples performs better than selecting fixed examples, indicating the value of example diversity in generation.
(4) Comparing \ding{175} and \ding{176}, we see that our proposed retrieval-based selection from a mixed pool performs comparably to selecting examples from a in-domain pool (which requires prior knowledge), indicating the effectiveness of our retrieval-based selection technique.
This result suggests that equipped with this technique, our proposed FedIT-U2S framework can be automatically deployed in various domains without much prior knowledge.

\begin{table}[t]
\centering
\begin{tabular}{l|c|c}
\toprule
Experimental Setup & Bert Score & ROUGE-L \\
\midrule
Base Model & 0.0566 & 0.2380 \\
\ding{172} Random 0 + 3 (3 out-domain examples) & 0.0868 & 0.2211 \\
\ding{173} Random 1 + 2 (1 in-domain and 2 out-domain examples) & 0.1143 & 0.2426 \\
\ding{174} Fixed 3 + 0 (3 fixed in-domain examples) & 0.1774 & 0.2942 \\
\ding{175} Random 3 + 0 (3 randomly selected in-domain examples) & 0.2128 & 0.3054 \\
\textbf{\ding{176} Retrieval-based Selection from A Mixed Pool} & 0.2035 & 0.2994 \\
\bottomrule
\end{tabular}
\caption{Experiments on HotpotQA dataset for analysis of example selection during few-shot data generation. The results show that our proposed automated retrieval-based selection technique can achieves comparable performance compared to selecting in-domain examples (which requires prior knowledge).}
\label{tab:domain}
\end{table}

\begin{figure}[t]
    \centering
    \begin{subfigure}[b]{0.3\textwidth}
        \centering
        \includegraphics[width=\textwidth]{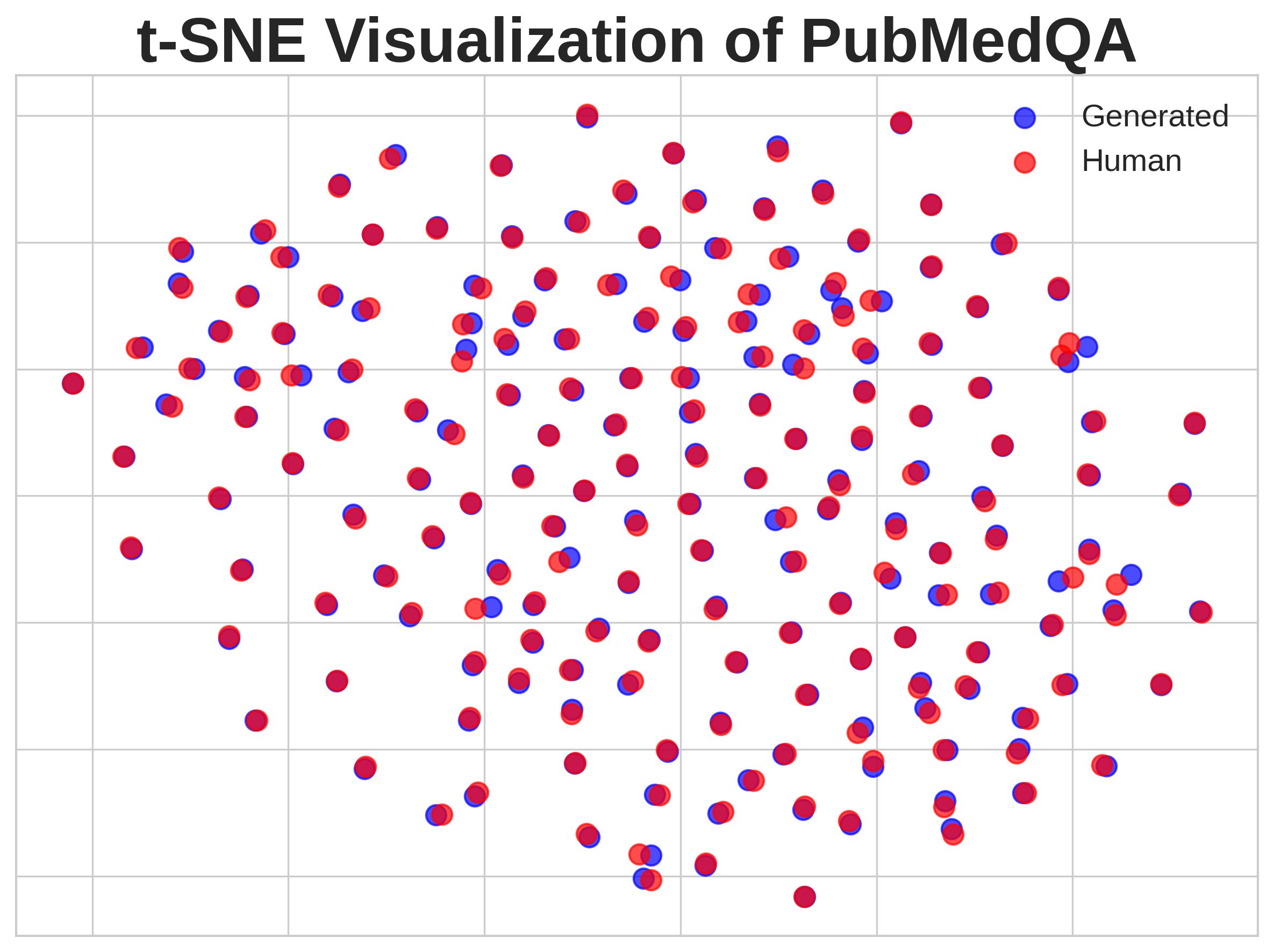}
        \caption{PubMedQA}
        \label{fig:tsne-med}
    \end{subfigure}
    \hfill
    \begin{subfigure}[b]{0.3\textwidth}
        \centering
        \includegraphics[width=\textwidth]{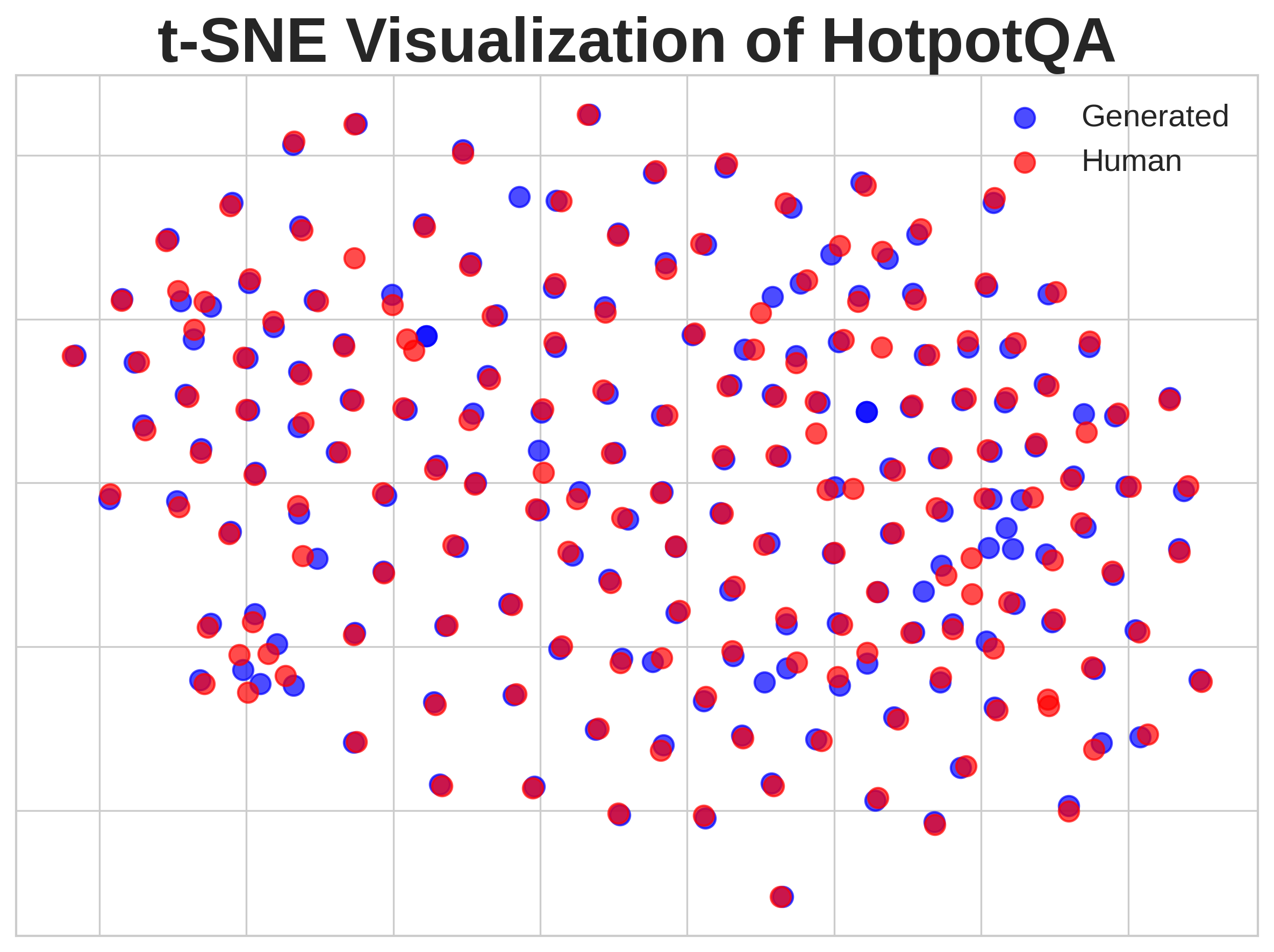}
        \caption{HotpotQA}
        \label{fig:tsne-knowledge}
    \end{subfigure}
    \hfill
    \begin{subfigure}[b]{0.3\textwidth}
        \centering
        \includegraphics[width=\textwidth]{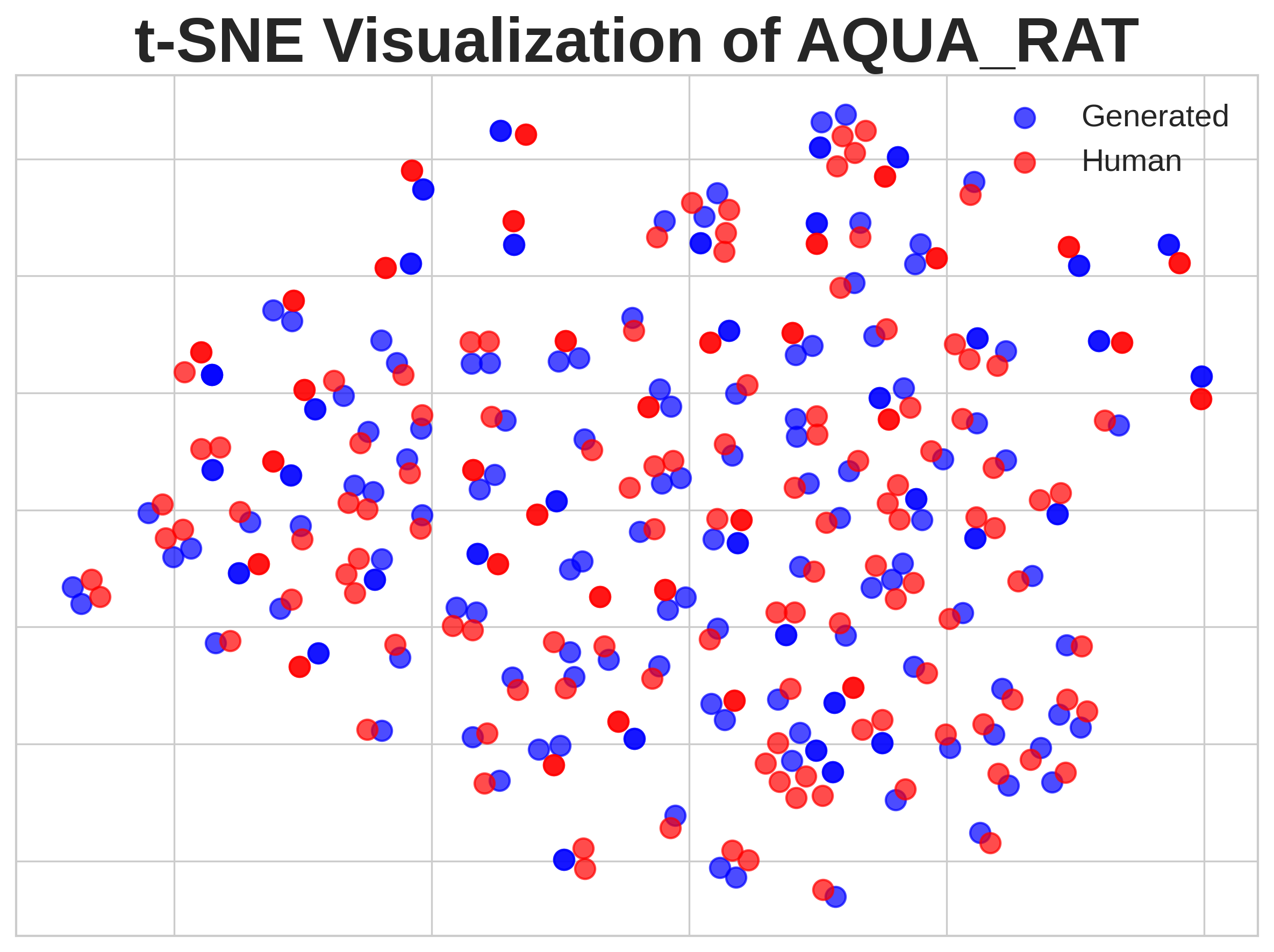}
        \caption{AQUA\_RAT}
        \label{fig:tsne-math}
    \end{subfigure}
    \caption{The t-SNE visualization of embeddings of instruction-response pairs in PubMedQA, HotpotQA and AQUA\_RAT. Blue dots represent generated data, while red dots represent human-annotated data. The close proximity of each pair of red and blue dots indicates that the generated data closely aligns with the human-annotated data. }
    \label{fig:tsne}
\end{figure}

\textbf{Comparisons of generated and human-annotated data.}
To better understand our method, we further analyze the characteristics of our generated data by comparing it with human-annotated data from two perspectives: embedding visualization and case studies.

(1) Embedding visualization: 
Here, we use t-SNE~\cite{van2008visualizing} to visualize the data points of generated and human-annotated data.
For each dataset, 200 generated and human-annotated sample pairs, sharing the same context, are selected.
The embeddings of the concatenated instruction and response texts are extracted via \textit{sentence-transformers}\footnote{\href{https://huggingface.co/sentence-transformers/all-MiniLM-L6-v2}{https://huggingface.co/sentence-transformers/all-MiniLM-L6-v2}} and mapped to a two-dimensional space via t-SNE.
The final 2D embeddings are plotted as shown in Figure \ref{fig:tsne}, where blue and red dots represent generated and human-annotated data respectively.
From the figure, we observe close proximity between the generated and human data points, indicating a high degree of alignment between the generated and human data across the datasets.

\begin{figure}[t]
    \centering
    \includegraphics[width=1.0\linewidth]{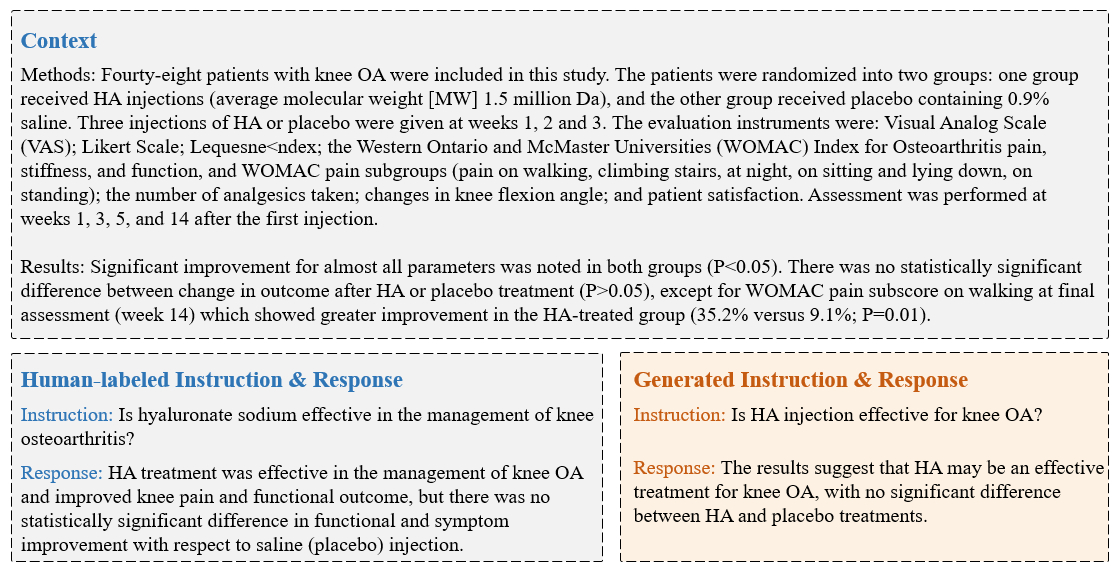}
    \caption{Example illustration.}
    \label{fig:case}
\end{figure}

(2) Case Study:  In Figure \ref{fig:case}, we show a specific example of generated data sample from PubMedQA.
The human-annotated data sample with the same context is also given for comparison.
Instructions of both samples ask about the effectiveness of HA injections in treating knee OA.
The generated response conveys a meaning similar to human-annotated response based on the context.

These two aspects of comparison demonstrate that our generated data is highly similar to the manually annotated data in both content and structure, reflecting the high quality of the generated data.

\section{Conclusions}

This paper proposes FedIT-U2S, which directly leverages clients' unstructured text data to achieve federated instruction tuning of large language models.
FedIT-U2S consists of two key steps: few-shot instruction-tuning data generation and federated instruction tuning on the generated data.
During data generation, for each unstructured data piece, a client firstly selects related examples via a retrieval-based example selection mechanism and then uses these examples for guiding the LLM to generate instruction-response pair based on the data piece.
A typical process of federated instruction tuning is then conducted based on the generated data.
Experiments on three domains (medicine, knowledge, and math) verify the effectiveness of our proposed FedIT-U2S.
Our method for the first time enables clients with unstructured data to be involved in the process of federated instruction tuning, which occupy a large proportion in practice and are underutilized previously.
We believe that this work can contribute to broadening the application scope of federated instruction tuning.

\clearpage

\medskip

\bibliographystyle{unsrt}
\bibliography{ref}

\appendix
\newpage

\section{Appendix}
\label{app}
\begin{lstlisting}[language=, frame=single, label=fewshot, caption=Few-shot prompt template]
Given the next [document], create a [question] and [answer] pair that 
are grounded in the main point of the document, don't add any 
additional information that is not in the document. The [question] is 
by an information-seeking user and the [answer] is provided by a 
helping AI Agent.

[document]: {The content of document 1}

### Response:
[question]: {The content of question 1}
[answer]: {The content of answer 1}

[document]: {The content of document 2}

### Response:
[question]: {The content of question 2}
[answer]: {The content of answer 2}

[document]: {The content document 3}

### Response:
[question]: {The content of question 3}
[answer]: {The content of answer 3}

[document]: {The content of the target text}

### Response:


\end{lstlisting}

\end{document}